\documentclass[letterpaper, 10 pt, conference]{ieeeconf}  %

\IEEEoverridecommandlockouts                              %

\usepackage{graphicx}
\usepackage{subfloat}
\usepackage[caption=false]{subfig}
\usepackage{tikz}

\usepackage{tabularx}
\usepackage{multirow}
\usepackage{booktabs}	%
 
\usepackage{color}

\usepackage{amssymb,amsmath}
\usepackage{amsfonts}

\usepackage{graphics} %
\usepackage{epsfig} %
\usepackage{mathptmx} %
\usepackage{times} %
\usepackage{amsmath} %
\usepackage{amssymb}  %

\usepackage{xifthen}
\usepackage{xparse}
\usepackage{subdepth}
\usepackage{leftidx}
\usepackage{bm}
\usepackage{amssymb,amsmath}
\usepackage{amsfonts}

\newcommand{\bbm}{\begin{bmatrix}}
	\newcommand{\ebm}{\end{bmatrix}}

\DeclareMathAlphabet{\mbf}{OT1}{ptm}{b}{n}
\newcommand{\mbs}[1]{{\bm{#1}}}

\newcommand{\mbsbar}[1]{{\overline{\boldsymbol{#1}}}}
\newcommand{\mbshat}[1]{{\hat{\boldsymbol{#1}}}}
\newcommand{\mbstilde}[1]{{\tilde{\boldsymbol{#1}}}}

\newcommand{\mbsdot}[1]{{\dot {\boldsymbol{#1}}}}

\newcommand{\mbfbar}[1]{{\overline{\mbf{#1}}}}
\newcommand{\mbfhat}[1]{{\hat{\mbf{#1}}}}
\newcommand{\mbftilde}[1]{{\tilde{\mbf{#1}}}}

\newcommand{\mbfdot}[1]{{\dot{\mbf{#1}}}}

\newcommand{\cframe}[1]{{\smash{\protect\underrightarrow{\mathcal{F}}_{#1}}}}

\DeclareMathAlphabet{\mathbfit}{OML}{cmm}{b}{it}
\newcommand{\homo}[1]{{\mathbfit{#1}}}
\newcommand{\mbfhbar}[1]{{\overline{\homo{#1}}}}

\newcommand{\mbfh}[1]{{\homo{#1}}}

\newcommand{\trans}[3]{\leftidx{_{#1}}{\mbf r}{\IfValueTF{#2}{_{#2#3\hspace{2pt}}}{}}} %

\newcommand{\vel}[3]{\leftidx{_{#1}}{\mbf v}{\IfValueTF{#2}{_{#2#3\hspace{2pt}}}{}}} %
\newcommand{\veltilde}[3]{\leftidx{_{#1}}{\mbftilde v}{\IfValueTF{#2}{_{#2#3\hspace{2pt}}}{}}} %
\newcommand{\velbar}[3]{\leftidx{_{#1}}{\mbfbar v}{\IfValueTF{#2}{_{#2#3\hspace{2pt}}}{}}} %
\newcommand{\velhat}[3]{\leftidx{_{#1}}{\mbfhat v}{\IfValueTF{#2}{_{#2#3\hspace{2pt}}}{}}} %
\newcommand{\veldot}[3]{\leftidx{_{#1}}{\mbfdot v}{\IfValueTF{#2}{_{#2#3\hspace{2pt}}}{}}} %

\newcommand{\myvec}[2]{\mbf{#1}_{#2}} %

\newcommand{\acc}[3]{\leftidx{_{#1}}{\mbf a}{\IfValueTF{#2}{_{#2#3\hspace{2pt}}}{}}} %
\newcommand{\acctilde}[3]{\leftidx{_{#1}}{\mbftilde a}{\IfValueTF{#2}{_{#2#3\hspace{2pt}}}{}}} %
\newcommand{\accbar}[3]{\leftidx{_{#1}}{\mbfbar a}{\IfValueTF{#2}{_{#2#3\hspace{2pt}}}{}}} %

\newcommand{\rotvel}[3]{\leftidx{_{#1}}{\mbs \omega}{\IfValueTF{#2}{_{#2#3\hspace{2pt}}}{}}} %
\newcommand{\rotveltilde}[3]{\leftidx{_{#1}}{\mbstilde \omega}{\IfValueTF{#2}{_{#2#3\hspace{2pt}}}{}}} %
\newcommand{\rotvelbar}[3]{\leftidx{_{#1}}{\mbsbar \omega}{\IfValueTF{#2}{_{#2#3\hspace{2pt}}}{}}} %
\newcommand{\rotvelhat}[3]{\leftidx{_{#1}}{\mbshat \omega}{\IfValueTF{#2}{_{#2#3\hspace{2pt}}}{}}} %
\newcommand{\rotveldot}[3]{\leftidx{_{#1}}{\mbsdot \omega}{\IfValueTF{#2}{_{#2#3\hspace{2pt}}}{}}} %

\newcommand{\Crot}[2]{\leftidx{}{\mbf C}{_{#1#2\hspace{2pt}}}} %
\newcommand{\Cbar}[2]{\leftidx{}{\mbfbar C}{_{#1#2\hspace{2pt}}}} %
\newcommand{\T}[2]{\leftidx{}{\mbfh T}{_{#1#2\hspace{2pt}}}} %
\newcommand{\Tbar}[2]{\leftidx{}{\mbfhbar T}{_{#1#2\hspace{2pt}}}} %
\newcommand{\pixel}[1]{{\mbfh u}{_{#1}}} %
\usepackage{multirow}
\usepackage{hyperref}

\newcommand{\paper}{paper}

\usepackage[capitalise,noabbrev]{cleveref}
\title{\LARGE \bf
MakeWay: Object-Aware Costmaps for Proactive Indoor Navigation Using LiDAR
}

\author{Binbin Xu$^{1}$, Allen Tao$^{1}$, Hugues Thomas$^{2}$, Jian Zhang$^{2}$, Timothy D. Barfoot$^{1}$
\thanks{1 The authors are with the Robotics Institute, University of Toronto, Toronto, ON M5S 1A1, Canada.
 	{\tt\small \{binbin.xu, allen.tao\}@robotics.utias.utoronto.ca,  
  tim.barfoot@utoronto.ca} 
 }
\thanks{2 The authors are with Apple, Cupertino, CA 95014 USA 
}
}

\begin{document}

\maketitle
\thispagestyle{empty}
\pagestyle{empty}

\begin{abstract}
In this paper, we introduce a LiDAR-based robot navigation system, based on novel object-aware affordance-based costmaps.
Utilizing a 3D object detection network, our system identifies objects of interest in LiDAR keyframes, refines their 3D poses with the Iterative Closest Point (ICP) algorithm, and tracks them via Kalman filters and the Hungarian algorithm for data association.
It then updates existing object poses with new associated detections and creates new object maps for unmatched detections. Using the maintained object-level mapping system, our system creates affordance-driven object costmaps for proactive collision avoidance in path planning.
Additionally, we address the scarcity of indoor semantic LiDAR data by introducing an automated labeling technique. This method utilizes a CAD model database for accurate ground-truth annotations, encompassing bounding boxes, positions, orientations, and point-wise semantics of each object in LiDAR sequences.
Our extensive evaluations, conducted in both simulated and real-world robot platforms, highlights the effectiveness of proactive object avoidance by using object affordance costmaps, enhancing robotic navigation safety and efficiency. The system can operate in real-time onboard and we intend to release our code and data for public use.

\end{abstract}

\section{Introduction}
Robot navigation has long been a core problem in many autonomous robotic applications. 
In a crowded environment, navigation tasks such as SLAM and path planning require careful treatments of static and dynamic elements, where semantic information can be exploited to enhance robustness and safety. For example, dynamic SLAM improves localization accuracy by differentiating between static and moving objects \cite{Xu:etal:ICRA2019} and path planing predicts pedestrian trajectories to prevent collisions \cite{Thomas:etal:ICRA2022, Salzmann:etal:RAL2023}. However, most existing works often rely on treating objects and humans as separate entities, resulting in a reactive approach to navigation that is based on simplified assumptions, such as constant velocity. In reality, anticipating the motion of all surrounding objects in dynamic environments is challenging and simply reacting to what is happening in the moment can be dangerous. For human beings, we tend to proactively anticipate higher risk areas based on the collision possibility of different objects, which is related to their semantic affordances. For example, a chair tends to be pushed and pulled by their users and thus would have higher collision risk near its front and back, compared to its sides. Therefore, current reactive approaches limit the potential for proactive, anticipatory movement in dynamic environments—a capability that is instinctive in human navigation.

To address these limitations, our research introduces a novel approach to robotic navigation in indoor settings. Based on object-level SLAM, we explore the integration of semantic information further into path planning, aiming to enable proactive collision avoidance. This approach is underpinned by the hypothesis: \textit{``Semantic information can be used for proactive risk avoidance to achieve safer and more efficient navigation"}. Recognizing the unique affordances of indoor environment objects, our method correlates path planning with surrounding semantic information, paving the way for a proactive navigation strategy.

\begin{figure}[t]
	\centering
	\includegraphics[width=0.9\linewidth]{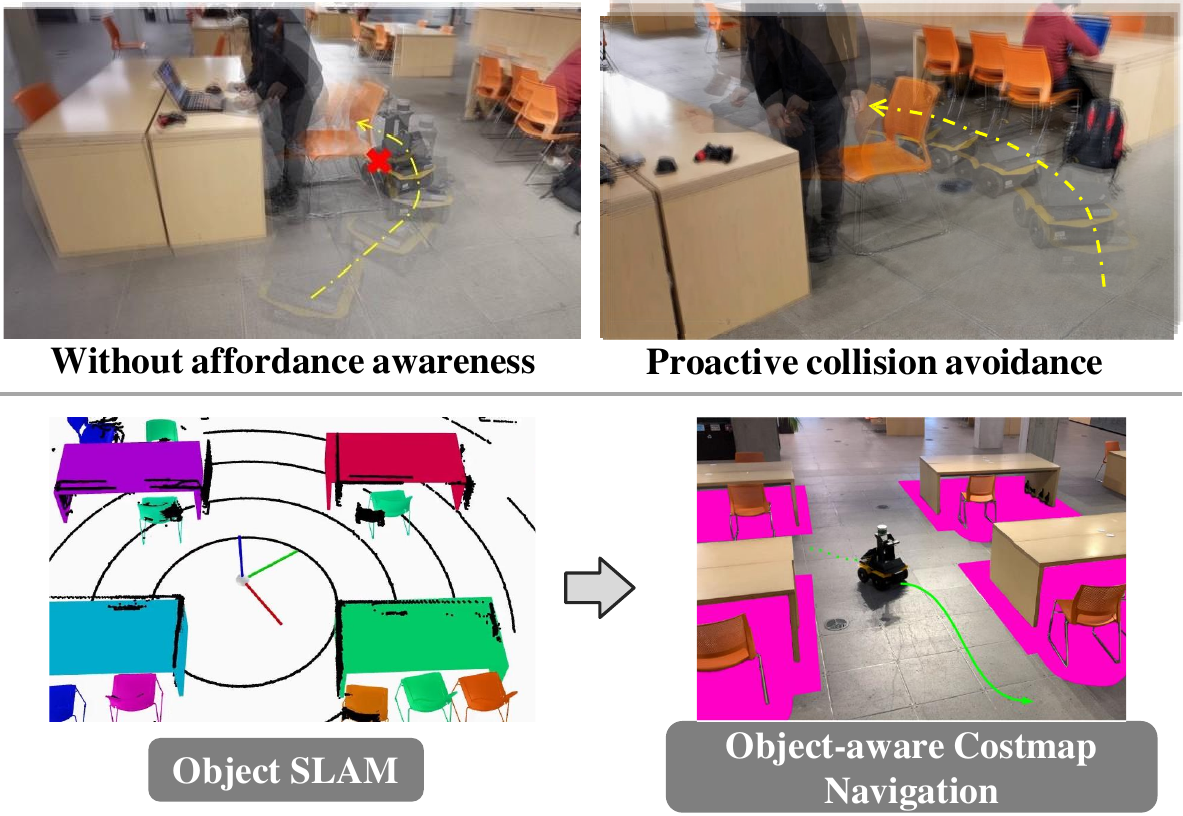}
	\caption{Incorporating semantic information for proactive collision avoidance, our system runs our object-level SLAM system and generates object-aware costmaps for indoor navigation.}
 \vspace{-1em}
	\label{fig:teaser}
\end{figure}

We present a LiDAR-based object-level SLAM system designed specifically for indoor navigation. Our system advances beyond traditional localization and mapping in dynamic settings by detecting, tracking, and reconstructing objects, then updating object-level costmaps with real-time pose and existence probability. This object-aware approach facilitates the creation of costmaps that incorporate object affordances, leading to enhanced safety and efficiency in navigation. The system's efficacy is validated quantitatively through extensive simulations in the Gazebo simulation environment and real-world tests using a Clearpath Jackal robot. Additionally, we validate our object costmap concept by comparing empirical data on typical object usage with our proposed affordance map.

Addressing the scarcity of public indoor LiDAR datasets with semantic annotations, we also introduce an automatic annotation system. Based on a multi-session SLAM setup \cite{Thomas:etal:ICRA2021}, this system can automatically annotate semantic points, instance points, object poses, and associate CAD models. Based on this automated annotation process, the semantic performance of our method can improve from session to session. %
The main contributions in this \paper~can be summarised as follows:
\begin{enumerate}
\item A LiDAR-based object-level system for indoor navigation, capable of detecting, tracking, and mapping objects, then leveraging this information for an object-aware costmap system;
\item An auto-labelling algorithm that annotates LiDAR sequences with ground-truth poses, bounding boxes, and point-wise semantic data using object 3D models;
\item Comprehensive experiments to evaluate the impact of object-level semantic information on semantic perception and navigation performance in both simulated environments and real-world scenarios.
\end{enumerate}

\section{Related Works}
\subsection{LiDAR-based Semantic SLAM}
LiDAR-based semantic SLAM, primarily focused on outdoor environments, has seen significant developments but remains underexplored for indoor applications. Pioneering work in this area, such as SegMap \cite{Dube:etal:IJRR2020, Cao:etal:IROS2021}, has demonstrated the utility of segmenting 3D point clouds into semantic categories. However, these methods typically concentrate on broader categories such as vehicles and buildings, lacking the granularity needed for intricate indoor environments. Subsequent studies, including those by Zhou et al. \cite{Zhou:etal:RAL2022} and Lusk and How \cite{Lusk:How:IROS2022}, have explored geometric detections such as lines and planes to improve data association and enhance localization accuracy. Yet, these approaches still fall short in creating detailed maps at the object level.

Our research fills this gap by introducing an indoor LiDAR-only object-level SLAM system. Our system not only maps objects but also creates object affordance costmaps for proactive collision avoidance. This advancement is especially crucial for navigating complex and dynamic indoor environments. Furthermore, our introduction of an automatic annotation system for indoor LiDAR data addresses a significant gap in the field. The lack of indoor datasets with semantic and instance annotations has restricted the development of comprehensive indoor navigation systems. Our system not only automates the annotation process but also evolves through unsupervised segmentation and model alignment, paving the way for lifelong learning in robotic navigation.

\subsection{Object-level SLAM}
Moving from semantic mapping to object mapping, object-level SLAM has made considerable advancements in environmental perception and mapping. Early works, such as the one by Wang et al. \cite{Wang:etal:ICRA2003}, laid the groundwork for object detection and tracking in unknown environments. Recent innovations in dense and dynamic SLAM, which integrate deep learning and visual data, have led to significant improvements through the development of more sophisticated segmentation and tracking algorithms~\cite{Xu:etal:ICRA2019, Rosinol:etal:ICRA2020}, while also enabling the maintenance of a spatial-temporal metric-semantic map~\cite{Schmid:etal:RSS2024}. Our method diverges significantly from these existing systems by not only maintaining object-level maps but also incorporating their affordances for indoor navigation where environmental dynamics are less predictable. We also deploy our object-level SLAM system on a real-world robot to showcase its benefits of scene understanding for indoor navigation and risk avoidance.
\subsection{Semantic Navigation}
The evolution of dense and semantic SLAM has spurred the inclusion of semantic data in navigation tasks. Recent studies have employed various approaches, such as using Bayesian multi-class semantic mapping for autonomous exploration \cite{Asgharivaskasi:Atanasov:ICRA2021}, including local object detection into the exploration utility function for actively searching and reconstructing objects of interests in unknown environments \cite{Papatheodorou:etal:ICRA2023}, and using reinforcement learning for semantic object exploration \cite{Chaplot:etal:NIPS2020}. While these methods enhance trajectory computation and environmental interaction, they often rely on static semantic maps and do not fully account for the dynamic nature of indoor environments.

Semantic information has been integrated into occupancy costmaps for path planning, particularly for collision avoidance in indoor navigation \cite{Lu:etal:IROS2014}. For instance, \cite{Pierson:etal:ICRA2019} predicts a Dynamic Risk Density from the occupancy density and velocity field of the environment based on object tracking and constant-velocity assumptions. Recent deep learning approaches, such as LSTM-based frameworks~\cite{Schreiber:etal:ICRA2020} and feedforward architectures~\cite{Thomas:etal:ICRA2022}, predict future environments using Dynamic Occupancy Grid Maps. Additionally, heuristic affordance safety zones based on humans' ages are manually designed around people for safety navigation in crowded environments \cite{Samsani:Muhammad:RAL2021}. However, these approaches primarily focus on \textit{reactive} navigation, predicting movements only within a limited timeframe (typically a few seconds) and often failing to anticipate sudden motion changes or delays in system inference. 

In contrast, our approach emphasizes \textit{proactive} navigation. Instead of solely reacting to immediate environmental changes, our method creates affordance collision risk areas directly from object occupancy maps. This object-centric approach, as opposed to prevalent point-centric models, can leverage completed object geometry to predict an occupancy map in unseen locations and uses semantic-dependent affordance to populate an instance-specific costmap in the scene.

\section{Method}
\label{sec:method}
\begin{figure}[tbp]
	\centering
	\includegraphics[width=\linewidth]{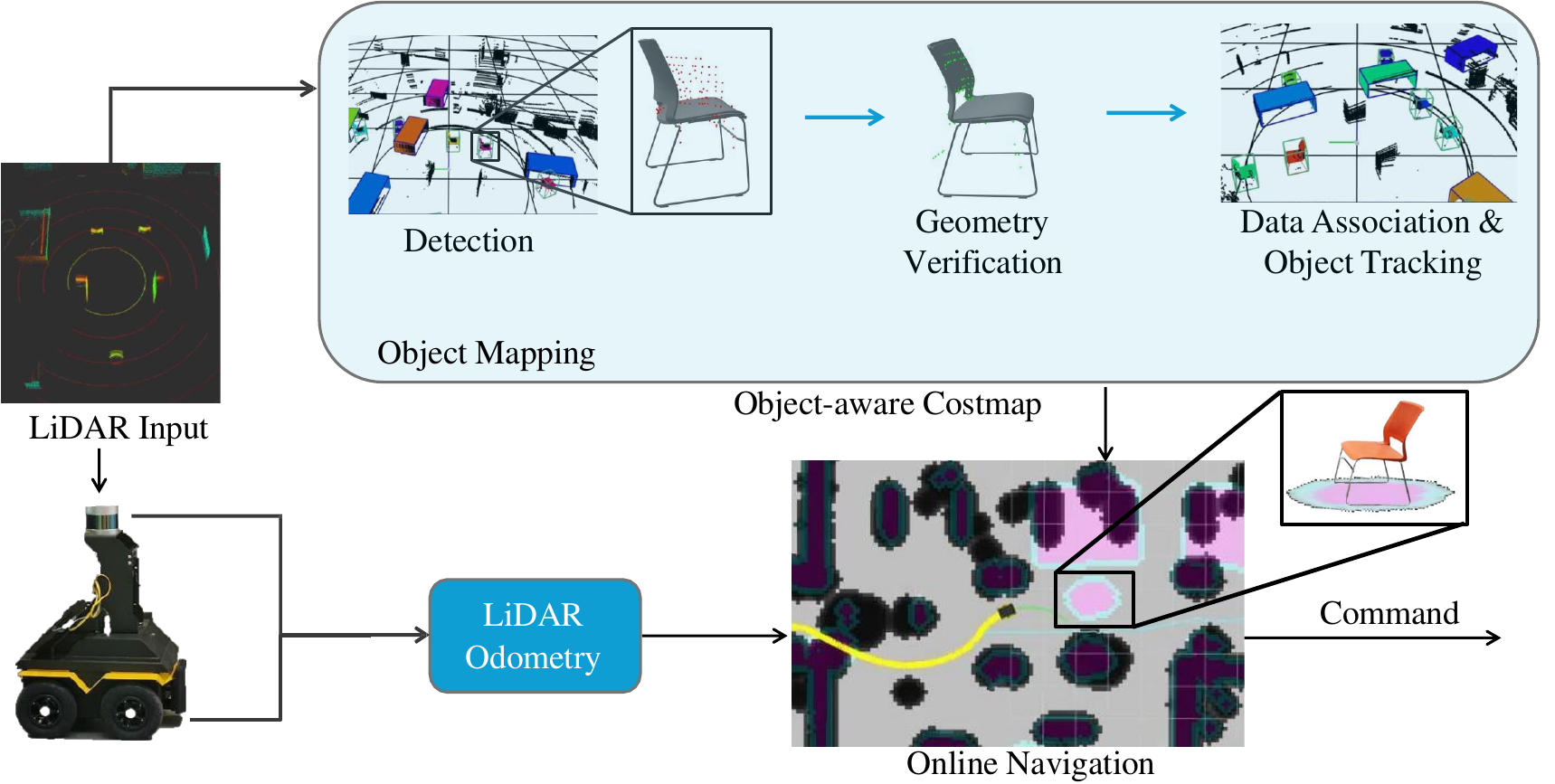}
	\caption{System architecture diagram. LiDAR input is consumed to produce 3D bounding boxes for chairs and tables. Their poses are then refined with ICP using their corresponding CAD models, and tracked across multiple frames. Next, our object-aware costmaps are attached to the tracked objects, which the planner uses to achieve object-aware proactive navigation.}
	\label{fig:pipeline}
\end{figure} 

\subsection{System Overview}
This work aims to explore the potential advantages of object awareness in robotic indoor navigation. Essentially, our robotic agent is designed to not only map its environment but also to identify objects, construct object-centric maps, and utilize these maps for safer navigation planning. Figure \cref{fig:pipeline} illustrates the workflow of our proposed system, which encompasses five components: odometry, detection, geometry verification, object tracking, and online navigation.

In our notation, we represent a coordinate frame as $\cframe{A}$. The homogeneous transformation matrix from $\cframe{B}$ to $\cframe{A}$ is expressed as $\T{A}{B}$, composed of its rotation $\Crot{A}{B}$ and translation $\myvec{r}{AB}$. A vector $\mbf a$ expressed in $\cframe{A}$ is $\myvec{a}{A}$. 

Our process begins with each incoming LiDAR frame, where we acquire the LiDAR point cloud $\myvec{p}{S}$ in the current sensor frame $\cframe{S}$. An existing point-based LiDAR-SLAM \cite{Thomas:etal:ICRA2021, Deschaud:ICRA2018} is run on each frame by fusing LiDAR measurements with robot odometry to estimate the sensor pose, $\T{W}{S}$, with respect to the map. To densify perception and reduce network detection latency, a sliding-window approach accumulates LiDAR points onto the latest keyframe using the estimated transformations before sending them to the object detection network. The object detection network takes the accumulated point cloud and predicts a group of object detection results $\myvec{x}{S}$ in the sensor frame. It is composed of the object center $\myvec{r}{S}\in \mathbb{R}^3$, bounding box scale $\myvec{b}{S}\in \mathbb{R}^3$, yaw orientation $\theta_{O}$, semantic label $s_l$, and detection confidence $s_c$:
\begin{align}
\label{eq: obj_det}
\myvec{x}{S} = \left\{ \myvec{r}{S}, \myvec{b}{S}, \theta_{O}, s_l, s_c \right\}.
\end{align}
The object center, scale, and yaw are further refined during geometric verification using known CAD models for objects. Then, object tracking creates data associations between the detections and object maps and continuously updates these maps using the associated detection results. For each unmatched object detection, we create an object model represented in its own object coordinate frame, $\cframe{O}$. The object costmaps (based on the object mapping) are subsequently integrated into the path-planning process.

\subsection{Object Mapping}
\noindent \textbf{Object Detection:~} 
Our system utilizes a sliding-window approach to accumulate LiDAR point clouds, which are then fed into an object-detection network at each keyframe. We have adapted the 3D object-detection framework from CAGroup3D \cite{Wang:etal:NIPS2022}, tailoring it for LiDAR detection to output a group of $\myvec{x}{S}$. Originally designed for axis-aligned, coloured point cloud input from RGB-D sensors (such as \cite{Dai:etal:CVPR2017}), it was retrained to accommodate colourless point cloud data from both LiDAR sensors by using mixed trained data from ScanNet \cite{Dai:etal:CVPR2017} and our annotated LiDAR data following the procedure in \cref{sec:annotation} to predict object orientations in the sensor coordinates.

\noindent \textbf{Ground Detection and Point Cloud Filtering:~}
To enhance network inference efficiency, we first detect the ground plane using a RANSAC-based plane-detection algorithm \cite{Fischler:Bolles:ACM1981}. This step, performed during the initialization stage, assumes that the ground plane contains more points than any other plane, a reasonable assumption for indoor mobile robots. By identifying the ground plane, we filter out ground points from each keyframe's accumulated point cloud before network processing. Additionally, we remove ceiling points based on their distance to the ground plane, focusing on objects that directly impact navigation.

\noindent \textbf{Geometry Verification:~} 
State-of-the-art 3D object detection algorithms, including the CAGroup3D model we adapted \cite{Wang:etal:NIPS2022}, typically output detections in bounding-box format. These models are trained using a metric known as rotated Intersection-Over-Union (rIOU), which compares the bounding box proposals to annotated ground-truth boxes. However, our experiments revealed a crucial limitation: high rIOU accuracy does not necessarily equate to accurate detection of object orientation. In cases of ambiguity between an object's width and length, the detected orientation may significantly differ from the actual orientation—by as much as $\frac{\pi}{2}$ or even in the complete opposite direction—while still maintaining high rIOU performance. Accurate orientation is vital for our object mapping and navigation phases, necessitating further optimization of the object pose.
 
To refine the predicted object pose, $\Tbar{S}{O}$ (derived from network outputs $\myvec{r}{S}$, $\myvec{b}{S}$, and $\theta_{O}$), we employ a customized Iterative Closest Point (ICP) algorithm. This process involves solving for a correction $\delta \T{}{}$ using the following ICP error:

\begin{align}
\label{eq: icp-residual}
\begin{split}
e_{ICP}(\delta \T{}{}) = & \left({\Cbar{S}{O} \myvec{n}{O}[\pixel{M}]}\right)^\intercal \cdot \\
                           & \left( \Tbar{S}{O} \myvec{p}{O} [\pixel{M}] - \delta \T{}{} \myvec{p}{S} [\pixel{S}]\right),
\end{split}
\end{align}

\noindent where $\myvec{p}{O}$ and $\myvec{n}{O}$ represent 3D voxel positions and normal vectors densely sampled from the object model, respectively, and $\myvec{p}{S}$ denotes the LiDAR point cloud within the detected bounding box. The correspondences, $\pixel{M}$ and $\pixel{S}$, are established through a hybrid Nearest-Neighbour Search \cite{Muja:Lowe:VISAPP}. The corresponding object CAD model can be retrieved from previous mapping sessions \cite{Thomas:etal:ICRA2021} or an a priori CAD model database. Given that objects are assumed to be on the ground, we constrain rotations about the $x$ and $y$ axes and focus on optimizing pose along the $z$-axis and translations. This approach expands the convergence basin, enabling more accurate pose correction. We apply a Cauchy robust loss function to the ICP residual and minimize it using a Gauss-Newton method in a three-level coarse-to-fine scheme.

For initialization, we calculate the geometric residual between the model and the LiDAR scan for four orientation candidates: ${\theta_{O}, \theta_{O}+\frac{\pi}{2}, \theta_{O}+\pi, \theta_{O}+\frac{3}{2}\pi}$. The orientation with the least error is selected as the starting point for ICP registration, addressing potential symmetric ambiguity and ensuring a more reliable initialization. This customized approach to CAD model fitting significantly enhances accuracy and robustness. If the final residual exceeds a predefined threshold, $e_{\rm max}$, the detection is rejected. Otherwise, we accept the detection and calculate its existence probability as
\begin{align}
\label{eq: exist}
p_{\rm exist} = 1 - \frac{1}{2}(e_{\rm icp} - e_{\rm min}) / (e_{\rm max} - e_{\rm min}).
\end{align}

\noindent \textbf{Data association:~} 
On each frame, our system associates new object detections with existing object models by calculating the 3D Generalized-Intersection-over-Union (GIoU) \cite{Rezatofighi:etal:CVPR2019} between each detection and model pair within the valid range of the sensor. Once a detection is associated with a model, we use 3D Kalman filters, based on AB3DMOT \cite{Weng:etal:IROS2020}, to carry forward the object's information from the previous keyframe to the current frame. The state vector of each object model is continuously updated using the 3D Kalman filter. This vector includes the object's position $\myvec{r}{S}$, orientation $\theta_{O}$, scale $\myvec{b}{S}$, and velocity $\myvec{v}{S}$. The object-detection network can provide a probability distribution $p(s_o|I_k)$ over the classes given the LiDAR keyframe $I_k$. Following \cite{Xu:etal:ICRA2019}, the semantic class probability distribution of each object model is updated through weighted averaging, refining its semantic label $s_l$ and detection confidence $s_c$:
\begin{align}
\label{eq: sem_fusion}
p(s_o|I_1, ..., I_k) &= \frac{1}{k}\sum_{i=1}^{k} p( s_o|I_i), \\
s_l &= \arg\max_{s_c} p(s_o|I_1, ..., I_k) \\
s_c = P(s_l|I_1, ..., I_k) &= \max_{s_c} p(s_o|I_1, ..., I_k)
\end{align}
The existence probability of each object is also updated using weighted averaging. This process helps refine the model's status by integrating new detection data.

New object detections that cannot be matched with existing models will initialize new object models. These models are centered around the object's position, with their coordinates relative to the world frame determined by the initial pose on this frame. Existing models that find no associated detections are set to zero existence probability on this frame. To address object inter-occlusion in cluttered indoor environments, we employ raycasting towards surrounding objects. This technique assesses whether objects are occluded by others, based on Axis-Aligned Bounding Boxes (AABB) intersections between rays and the bounding boxes of nearby objects. If occlusion is detected, we do not penalize the existence probability of the occluded objects. Objects that are continuously undetected over several frames, or whose existence probability falls below 0.5, are removed from the object map database. This approach ensures that our object map keeps reliable and accurate object detections and removes false positive ones.

\subsection{Online Navigation}
Once the sensor pose, $\T{W}{S}$, and the poses of object models, $\T{W}{O}$, are estimated, we generate and update object-level costmaps for online navigation. These costmaps are integrated into the ROS navigation stack \cite{Marder:etal:ICRA2010}, consisting of a 2D background costmap and multiple foreground object-centric costmaps.

\begin{figure}[tb]
	\centering
        \includegraphics[width=\linewidth]{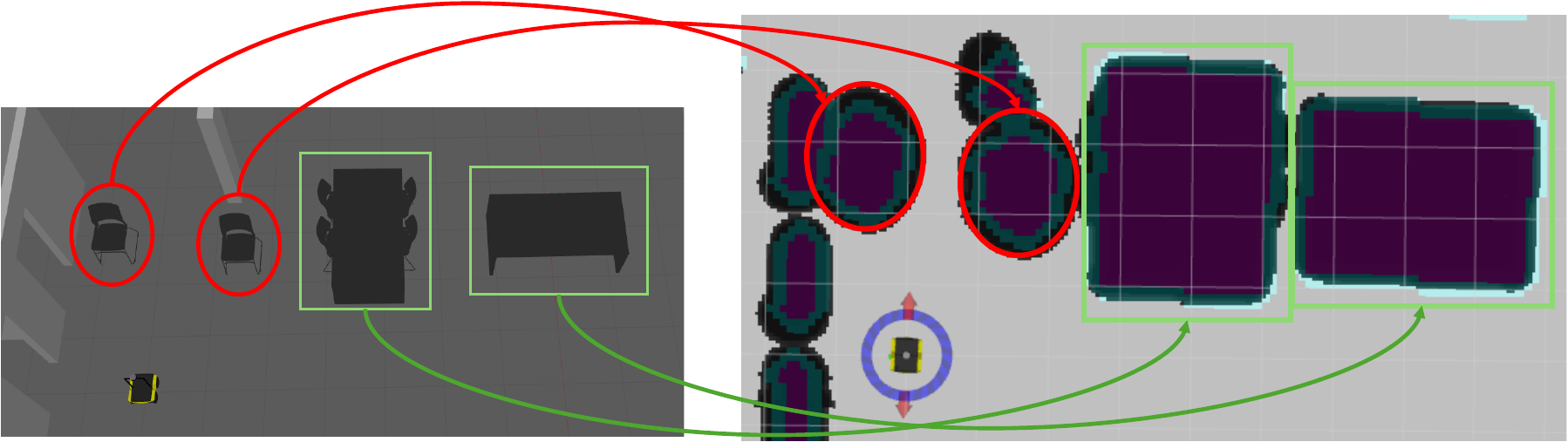}
	\caption{Object-aware costmaps. Our object-level dense mapping is on the left, and the right is the obtained costmap. Based on the affordance of chairs, we assign our elliptical cost aligned with the chairs' detected orientation.}
	\label{fig:costmap}
\end{figure}

To enable proactive collision avoidance, our costmaps are designed for different semantic classes, taking into account their unique affordances. For instance, costmaps for chairs account for their high likelihood to be slid backwards or forwards, forming an elliptical shape aligned with the chair's orientation. Similarly, costmaps for tables are rectangular with additional padding on the sides to account for typical human sitting positions; the tables we used did not allow for sitting at the ends. These costmaps feature higher costs at the center, gradually decreasing outward, as illustrated in \cref{fig:costmap}. For non-semantic background objects, a standard background costmap is created based on existing mapping \cite{Grisetti:etal:TRO2007}. The ROS 2D costmap package processes these maps by merging the background obstacle map with the foreground object costmaps. This combined map is then utilized for navigation planning. The estimated robot pose from odometry, along with a defined final goal, guides our path planning. We employ A* \cite{Hart:etal:Astar} for global path planning and the Dynamic Window Approach (DWA) \cite{Fox:etal:RAM1997} for local trajectory planning, ensuring feasible and safe navigation.

\subsection{Data Annotation}
\label{sec:annotation}
To address the scarcity of annotated indoor LiDAR semantic datasets, we developed an annotation pipeline to label 3D point cloud with semantic, instance, bounding box, and pose annotations.
Our annotation pipeline is illustrated in \cref{fig:annotation} and is composed of following steps:

\begin{figure}[tb]
	\centering
	\includegraphics[width=\linewidth]{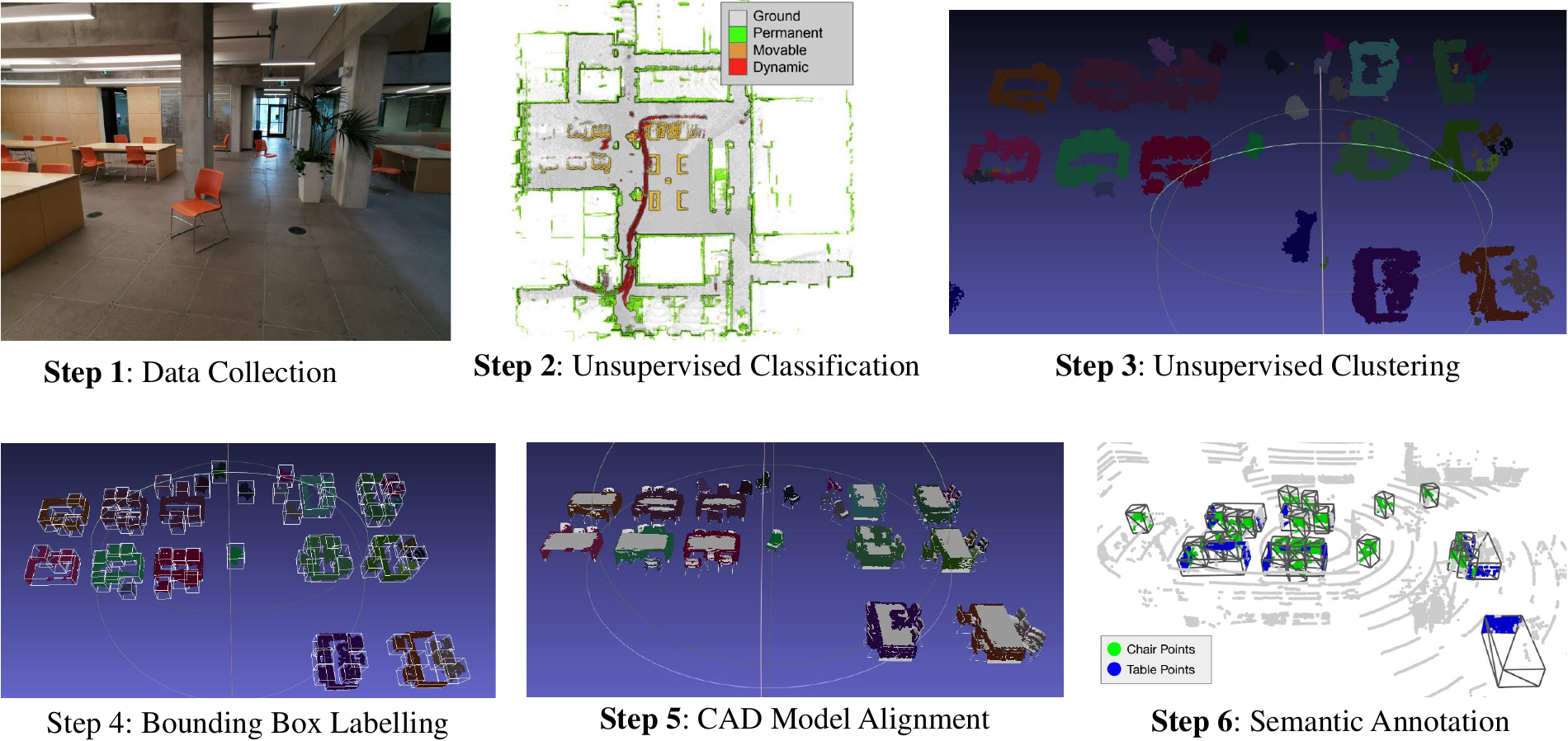}
	\caption{Our six-step data annotation pipeline. LiDAR point clouds are collected by the robot. The points are then classified, keeping the Movable points corresponding to occasionally moved objects (i.e., tables and chairs). Objects are then isolated using clustering, and rough 3D boxes are drawn around them, providing initializations for ICP-based CAD alignment. Finally, all Movable points are semantically classified as tables or chairs.}
	\label{fig:annotation}
\end{figure}

\begin{enumerate}
    \item Data Collection: 
    We gather raw LiDAR data using a Velodyne Ultra Puck LiDAR sensor mounted on a Clearpath Jackal robot. We capture data from diverse trajectories and changing indoor configurations.
    \item Unsupervised Classification: 
    Utilizing multi-session SLAM and ray tracing probabilities \cite{Thomas:etal:ICRA2021}, the collected data points are categorized into four labels: Ground, Permanent, Movable, and Dynamic, based on their movable probability. The Movable label is particularly significant, identifying objects likely to be repositioned by human interaction, such as furniture, and demanding heightened attention for collision avoidance.
    \item Unsupervised Clustering: 
    Focusing on Movable labeled points, we employ the DBScan algorithm \cite{Schubert:etal:DBSCAN} to segment them into discrete objects. 
    \item Bounding Box Labelling:
    We compute the oriented bounding box based on the PCA of each clustered point cloud. Manual intervention is occasionally needed, especially when objects are in close proximity and challenge the clustering algorithm. In such instances, users manually draw 3D bounding boxes around the missing objects. Notably, this is the only manual component of our data annotation pipeline and could be mitigated by leveraging object detectors trained on past sessions.
    \item CAD Model Alignment: 
    Each segmented point cloud is then automatically aligned with a corresponding CAD model using multi-scale ICP (as detailed in Eq. \ref{eq: icp-residual}). Then we retain only the rotation around the $z$-axis from the estimated transformation. The 3D bounding boxes for objects are extracted by computing the dimensions of the aligned CAD models.
    \item Semantic Annotation:
    Finally, we assign semantic and instance labels to each point within the object point clouds, based on their proximity to the aligned CAD models. Points that fall beyond a set distance threshold from the CAD model surface are considered outliers and excluded. This step ensures accurate and precise semantic classification of each point.
\end{enumerate}

\subsection{Implementation Details}
To train our 3D detection network for LiDAR inputs, we utilized a diverse dataset comprising both ScanNet \cite{Dai:etal:CVPR2017} and our custom annotated data (as outlined in \cref{sec:annotation}) to ensure the diversity of semantic categories and point cloud densities. Our training set includes 1201 point-cloud samples from ScanNet and 816 from our dataset. To bridge the gap between the RGB-D point clouds in ScanNet and our sparse, colorless LiDAR data, we removed color information from ScanNet samples and applied randomized subsampling and rotations in data augmentation. The network architecture, inspired by \cite{Wang:etal:NIPS2022}, was trained for 20 epochs using the AdamW optimizer \cite{Loshchilov:Hutter:ICLR2019}, with an initial learning rate of 0.001 and weight decay of 0.0001.

For online navigation, we integrated our system within the ROS framework, with key components including odometry, object detection, model alignment, and navigation operating as separate ROS nodes. This parallelization allows for real-time localization, as odometry is processed with each frame. For object detection, we accumulate every 5 consecutive LiDAR frames to ensure that the detection network has a sufficiently dense point cloud for inference considering the LiDAR's update rate and network inference latency. To manage existing object maps and trackers, we utilize a hashtable. The detection module is implemented in PyTorch while the rest of the components are in C++ for efficiency.

\section{Experiments}
\subsection{Simulated Experiments}
\subsubsection{Experimental Setup}

\begin{figure}[tb]
	\centering
 \includegraphics[width=0.9\linewidth]{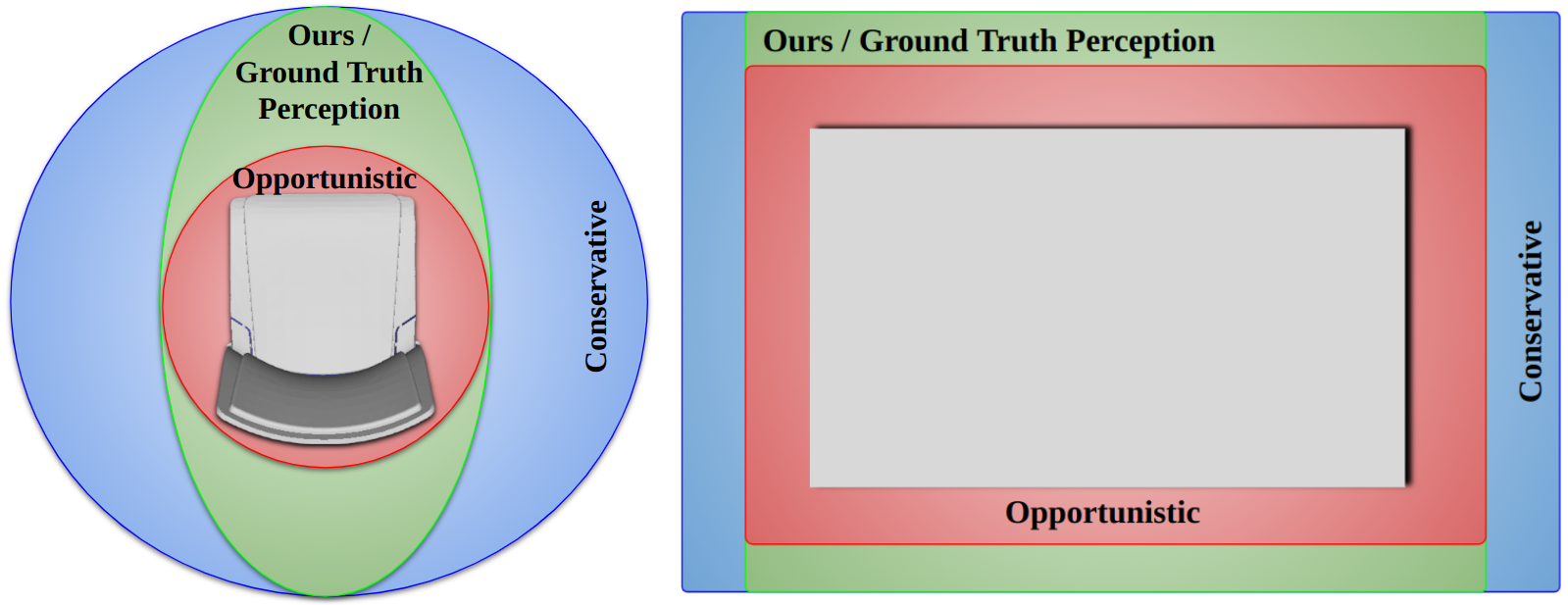}
	\caption{Chair (left) and table (right) costmaps for Experimental Setup.}
	\label{fig:costmaps}
\end{figure}

To investigate the effects of the object-aware costmap for navigation, we evaluate against two variants of a classic costmap-based navigation algorithm \cite{Lu:etal:IROS2014}  that are unaware of semantic information and two versions of our system in both simulations and real-world robot platforms. The chair and table costmaps used for each planner is visualized in Figure \ref{fig:costmaps}.
\begin{itemize}
    \item Conservative planner (\textit{Con}): prioritizes safety by maintaining an extra clearance distance from all obstacles to minimize collision risks - it does not factor in the orientation of objects and uniformly avoids risk across all directions.
    \item Opportunistic planner (\textit{Oppo}): Prioritizing efficiency, this model seeks the shortest travel times by keeping minimal clearance from obstacles, essentially treating all objects as static and unchangeable.
    \item Ours with Ground Truth Perception (\textit{GT-Perc}): Showing the efficacy of our costmap and navigation component, this version employs ground truth object detection in the scene to create affordance-informed costmaps.
    \item Ours (\textit{Ours}): a full system as described in \cref{sec:method}, integrating detection, geometric verification, object tracking, mapping, and costmap systems.

\end{itemize}

All planners' local costmaps include the static, obstacle, and inflation layers. The conservative one has an inflation radius being 3 times of the robot's radius. %
The opportunistic one has an inflation radius the same as the robot's radius. %
For fair comparisons, we maintain the same parameters for local and global planners in all experiments.

To quantitatively evaluate the navigation performance, we use the metrics of time taken to reach the goal $t_g$, distance taken to reach the goal $d_g$ and time spent in risky areas $t_r$.
The risk time is defined as the amount of time the robot spends in the defined affordance-based costmaps around each object.

We also evaluated the semantic perception performance by evaluating the mean average precision (mAP) and mean average recall (mAR) with different IoU thresholds, i.e., 0.25 and 0.50 as well as the accuracy of the estimated orientation for detected objects.

\subsection{Simulation}
\begin{figure}[t]
    \centering
    \includegraphics[width=1\linewidth]{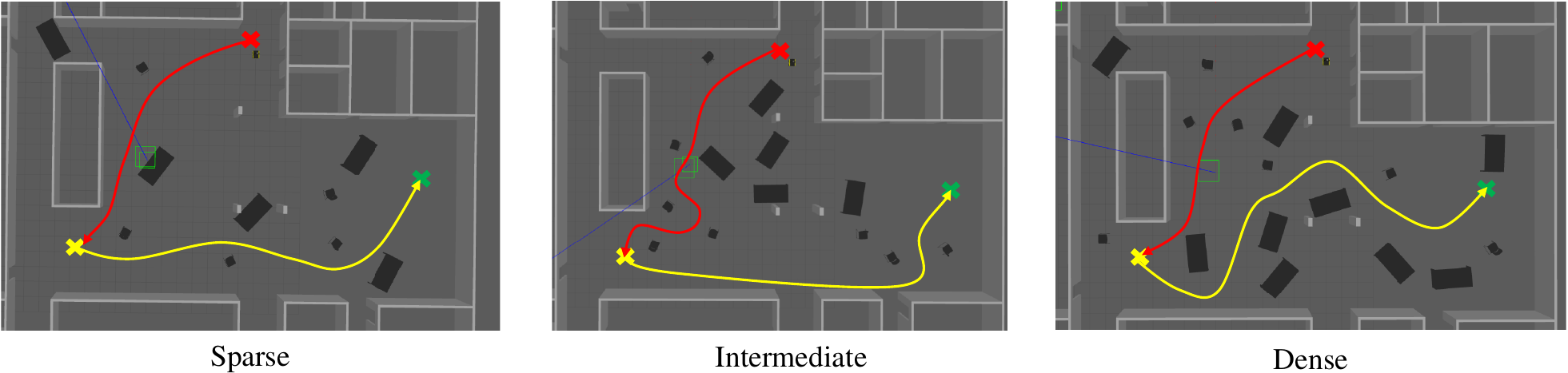}
    \caption{Examples of our simulation environments. The red cross $\textcolor{red}{\times}$ denotes the start, the yellow cross $\textcolor{yellow}{\times}$ denotes an intermediate waypoint, and the green cross $\textcolor{green}{\times}$ denotes the goal.}
    \label{fig:sim_maps}
\end{figure}

For a thorough and reproducible evaluation, we designed an indoor navigation simulation using Gazebo. We varied the environment's sparsity and complexity by randomizing the objects’ number and locations and created 25 different scenarios for robot navigation. 
Examples of density diversified simulation environment is shown in \cref{fig:sim_maps}.
Consistent parameters were maintained for both local and global planners across all tests. We measured navigation performance using metrics of total time to reach the goal, time spent in high-risk areas, and distance traveled to reach the destination.  The simulated Jackal uses only a LiDAR sensor, identical to our real world Jackal. In Gazebo, it drives at a top speed of 0.75 meters per second. 
All simulated experiments were run on a laptop with an Intel Core i7-12800HX CPU, 16 GB of memory and a 16GB NVIDIA A4500 Mobile GPU. They were run on Ubuntu 20.04 using ROS Noetic and compiled with GCC 9.4.0 using the O3 optimisation level. 

\begin{table}[t]
\scriptsize
\centering
\caption{Comparison of Different Methods in Simulations. We report sum/mean averaged by the success runs.}
\begin{tabular}{c|c|c|c|c}
\hline
\textbf{Methods} & \textbf{Total Time (s)} & \textbf{Risk Time (s)} & \textbf{Distance (m)} & \textbf{Success} \\ \hline
Con.       & 1006.04/55.89 & 0.00/0.00    & 761.76/42.32 & 18/25 \\ \hline
Oppo.      & 1280.65/51.23 & 78.80/3.15   & 976.41/39.06 & 25/25 \\ \hline
GT-Perc  & 1286.19/51.45 & 0.00/0.00    & 998.21/39.93 & 25/25 \\ \hline
Ours & 1314.05/52.56 & 12.99/0.52  & 1011.98/40.48 & 25/25 \\ \hline

\end{tabular}
\label{tab:sim_exp}
\end{table}

\cref{tab:sim_exp} summarizes a comparative analysis across different methods in all 25 runs in Gazebo simulations. Our systems, both \textit{Ours} and \textit{GT-Perc}, consistently outperformed the others, particularly in safety and the ability to navigate larger areas successfully. Their perfect success rates underline their reliability, essential in real-world applications where consistency is vital. The zero total risk time in \textit{GT-Perc} and the very low risk time in \textit{Ours} underscore the importance placed on safety in these systems. Notably, the safety measures in our systems scarcely compromise efficiency, as shown by their travel distances and success rates.

\begin{figure}[t]
     \centering
     \includegraphics[width=0.8\linewidth]{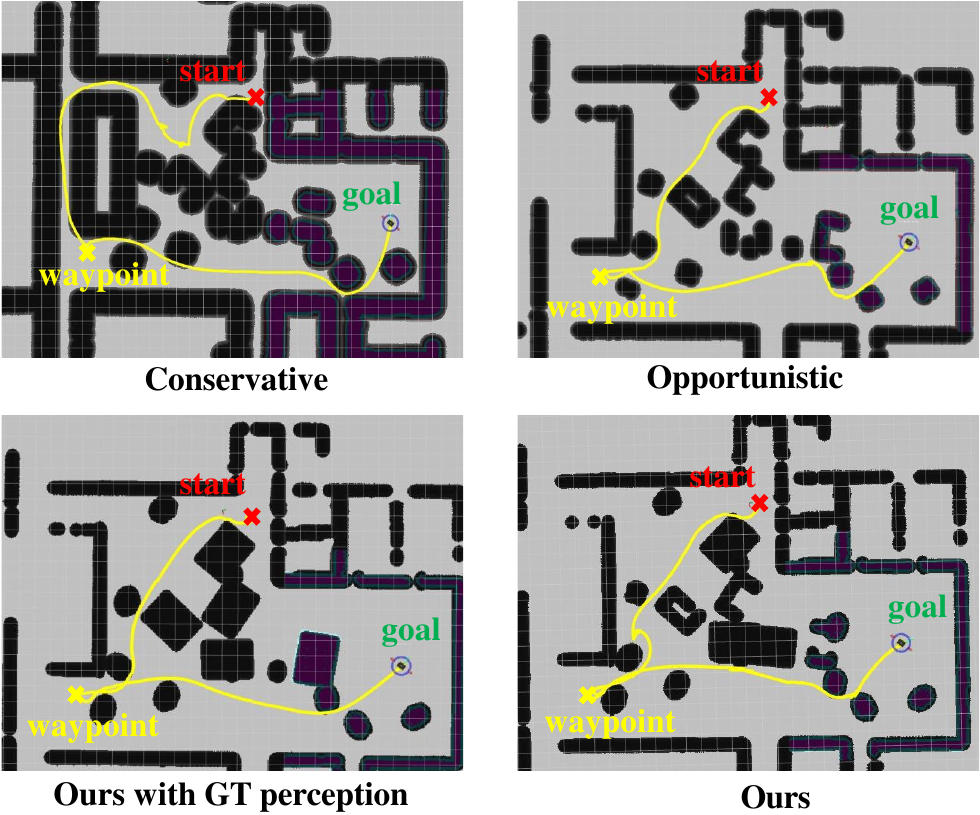}
     \caption{Visualization of different methods’ navigation routes in an intermediate dense scene.
}
     \label{fig:sim_qua}
 \end{figure}

The opportunistic planner (\textit{Oppo}) completes tasks in the least amount of time. However, it tends to take risks by traversing through potentially hazardous areas in its quest to shorten the path. This could be a concern in scenarios where safety is of paramount importance. 
On the other hand, the conservative planner (\textit{Con}) prioritizes safety by avoiding risky areas entirely, but this cautious approach often results in longer travel times. Besides, in some clustered environments, it may even fail to complete the assigned tours, as shown by its lower success rate. This suggests that while \textit{Con} may operate well in controlled or sparse environments, its performance to reach its goal is compromised and less reliable in more clustered environments. 
In contrast, our system proactively avoids high-risk areas, ensuring safety without compromising on operational goals. \cref{fig:sim_qua} illustrates the navigation routes driven by different navigation policies in an intermediate dense scene. 

\subsection{Real-world experiments}
\begin{figure}[t]
    \centering
    \includegraphics[width=\linewidth]{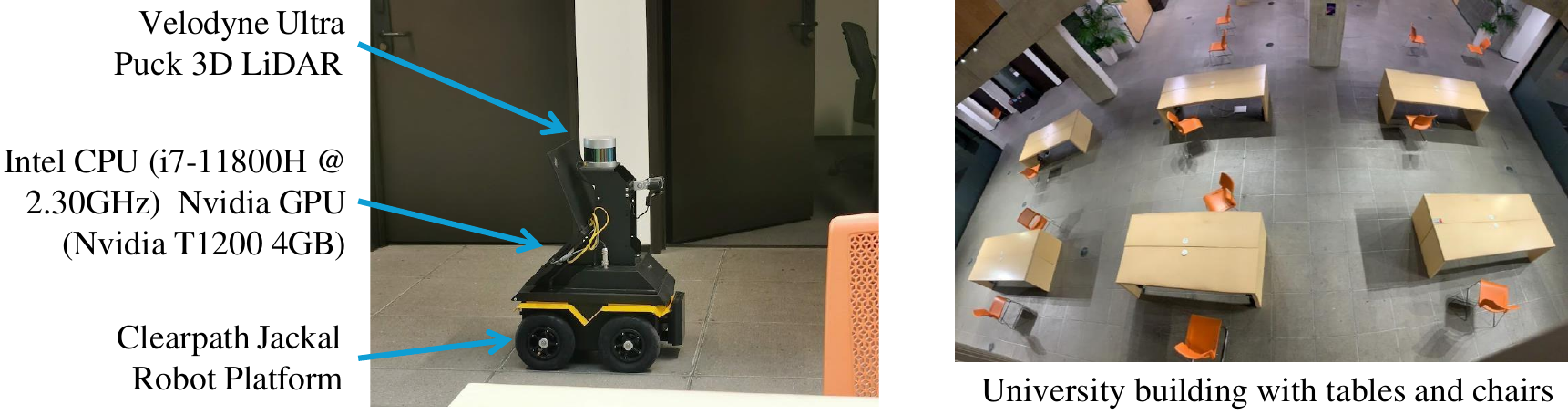}
    \caption{Real world indoor navigation setup}
    \label{fig:realworld}
\end{figure}

We further conducted our experiments in a real-world indoor study space at a university campus  
to showcase the feasibility of using the proposed approach onboard. The mobile robot used for the experiment is a Clearpath Jackal robot equipped with a Velodyne Ultra Puck LiDAR sensor. All processing is computed using an onboard computer with an Intel Core i7-11800H CPU, 16 GB of memory and an NVIDIA T1200 4GB GPU. The top speed of the Jackal is set as 0.75 m/s. The environment contains 12 tables and 22 chairs, as shown in \cref{fig:realworld}. We designed three different routes with different starting points and destinations that are feasible for all three tested methods to achieve. Using our auto-labeling system, we generated ground-truth data for object locations and orientations. We further included metrics for 3D object detection and orientation estimation accuracy.

\begin{figure}[t]
    \centering
    \includegraphics[width=1\linewidth]{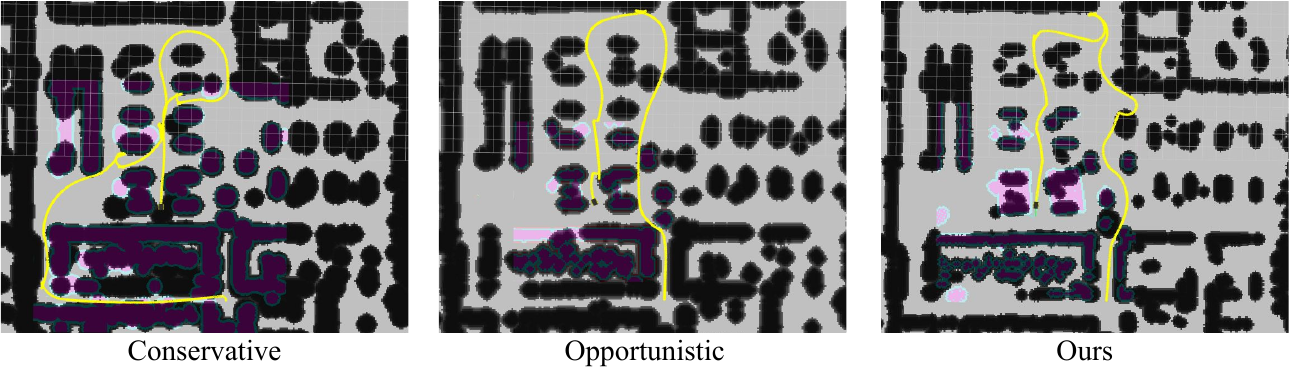}
    \caption{Visualizations of different methods’ navigation routes in a real world scene}
    \label{fig:real_qua}
\end{figure}

\begin{table}[t]
\centering
\scriptsize
\caption{Comparison of Different Methods in Real-World Runs}

\begin{tabular}{c|c|c|c}
\hline
\textbf{Methods} & \textbf{Total time (s)} & \textbf{Risk Time (s)} & \textbf{Distance (m)} \\ \hline
Con      & 584.103 & 7.327 & 197.725 \\ \hline
Oppo     & 390.216 & 31.377 & 147.377 \\ \hline
Ours & 478.087 & 7.496 & 165.975 \\ \hline
\end{tabular}
\label{tab:robot_navigation}
\end{table}

\cref{tab:robot_navigation} summarizes results from all three different real-world trials and \cref{fig:real_qua} visualizes the result from one of the test routes. These results highlight the advantages of our object-aware costmap, particularly in enhancing navigation safety and efficiency. For instance, \textit{Oppo} achieved the shortest total time but at a significantly higher total risk whereas \textit{Ours} offered a balance between efficiency and reduced risk exposure. Furthermore, our system's ability to navigate with a minimal risk profile, closely rivaling \textit{Con}'s risk time, highlights its proficiency in safe navigation while taking a shorter route thanks to less restrictive object-aware costmaps. This suggests that object-aware costmaps effectively inform navigation, allowing the system to proactively avoid high-risk areas while maintaining a competitive operational pace.

\begin{figure}[t]
    \centering
    \includegraphics[width=0.9\linewidth]{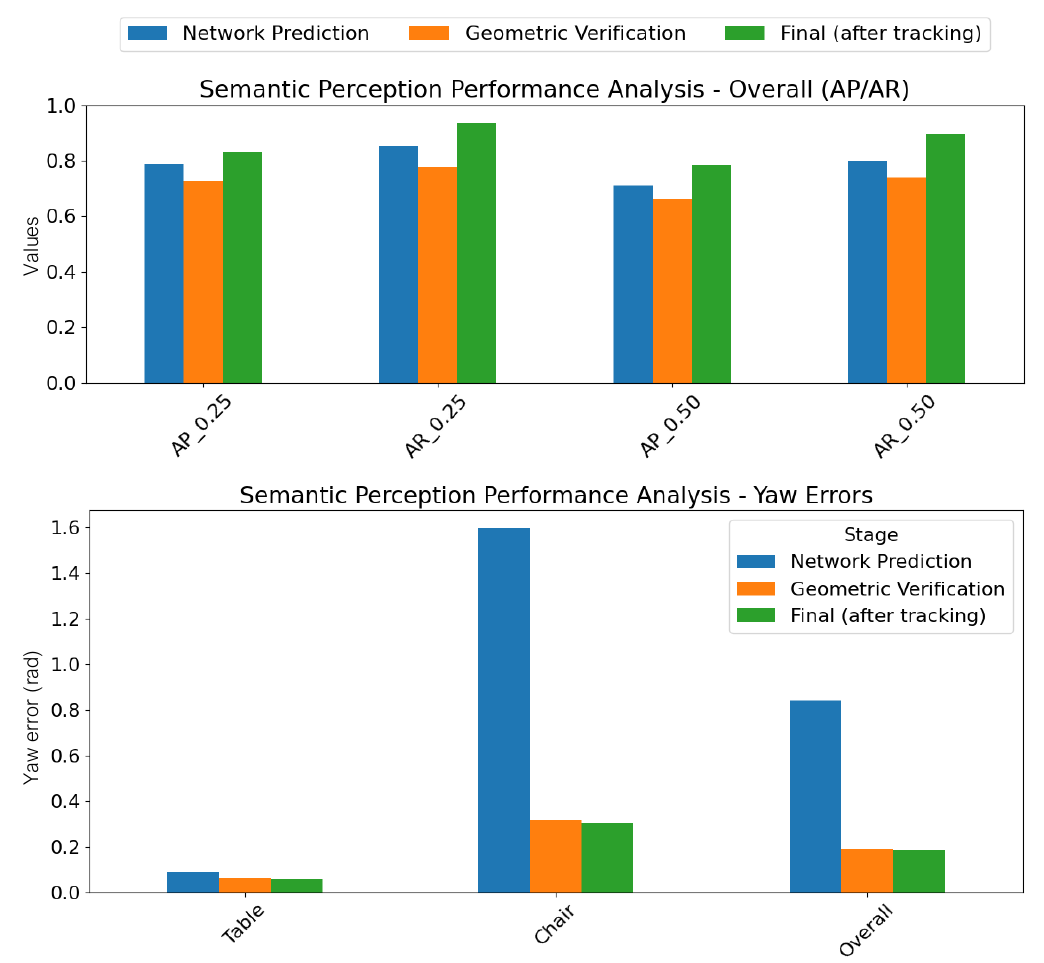}
    \caption{Ablation studies of different stages on 3D object detection and object orientation estimation in our object SLAM system.}
    \label{fig:real_sem}
\end{figure}

We further conducted ablation studies in the real-world experiments on the semantic performance of each stage in our mapping component. \cref{fig:real_sem} reports the results after each stage. We can see that with geometric verification, despite a slight drop in detection accuracies, orientation accuracy improved significantly. There is a trade-off between detection accuracy and orientation reliability. Nonetheless, geometric verification is essential for addressing challenges in scenarios with sparse point clouds or difficult viewpoints, which hinder effective alignment using ICP. Consequently, we selectively filtered detections to retain only those with confident orientation estimations. The final system configuration, incorporating tracking to utilize temporal measurements, enhanced both detection and orientation accuracy. This balanced improvement underscores the effectiveness of our approach in optimizing the system for accurate and reliable performance in complex real-world scenarios.

\subsection{Real-world affordance analysis}

\begin{figure}[t]
    \centering
    \includegraphics[width=1\linewidth]{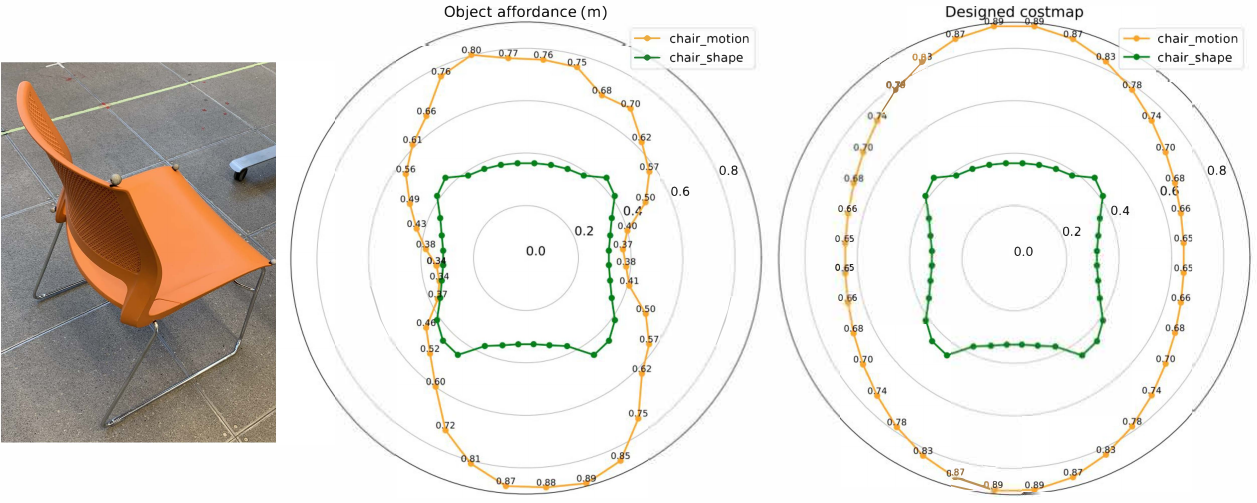}
    \caption{Object costmap analysis: from left to right: objects used in vicon data collection,  object (chair) affordance based on distribution of point cloud motion magnitudes within discrete motion direction bins, our designed costmap for the chair class.}
    \label{fig:real_affordnace}
\end{figure}

To verify our essential hypothesis on the affordance-driven costmap, we conducted a validation of our object costmap concept by analyzing the affordance map of objects used in our experiments. As demonstrated in the left image of \cref{fig:real_affordnace}, we established a test environment within a Vicon motion capture space. We attached Vicon markers to a chair's surface to capture multiple real-world sequences, observing its utilization in daily activities.

The collected data, showcasing ground-truth motion trajectories, revealed distinct movement patterns. Particularly noteworthy, as depicted in the central image of \cref{fig:real_affordnace}, was the motion pattern of the chair's point clouds in different directions. They predominantly exhibited extended motion along their forward and backward axes. This empirical observation aligns with our initially hypothesized design, as illustrated on the right side of \cref{fig:real_affordnace}. Here, we had intuitively opted for an elliptical object affordance costmap specific to the chair semantic class. The empirical data corroborated our heuristic approach, underscoring the effectiveness and relevance of our costmap design in real-world scenarios. This alignment between real-world usage patterns and our predictive costmap model not only validates our approach but also enhances our confidence in the practical application of these theoretical concepts in dynamic environments.

\subsection{Limitations}
As noticed in \cref{tab:robot_navigation}, our real-world experiments revealed some instances of risk time in both the Conservative Planner (Con) and Our Full System. This occurrence is attributed to two primary factors: imperfect detection and constraints of the global planner. For the first factor, our system, at times, fails to detect some objects in the scene. This issue is particularly prevalent in areas where objects are clustered, such as a chair tucked under a table. Such undetected objects contribute to the risk time, as the system may not adequately account for these obstacles in its navigation strategy. For the second factor, the A* algorithm \cite{Hart:etal:Astar} used for global planning does not always propose physically feasible routes for our robots. Consequently, the local planner is occasionally compelled to deviate from the planned path to reach the goal. These deviations, while necessary, introduce risk time as the robot navigates through less-optimized paths.

\section{Conclusions}
In this paper, we present a novel object-aware navigation system using LiDAR, showing how an awareness of surrounding objects can substantially improve both the safety and efficiency of robotic navigation. 
We have evaluated its effectiveness through simulations and real-world experiments and have demonstrated that it can run efficiently on real-world robotic platforms. 

While our object-level costmaps are effective, they are currently based on manual designs, derived from empirical observations of object behaviors and affordances. Looking forward, we will focus on developing a data-driven approach to refine our object affordance costmaps and enhance the object-aware navigation system.

\bibliographystyle{IEEEtran}
\bibliography{robotvision}

\end{document}


\maketitle
\thispagestyle{empty}
\pagestyle{empty}

The learnable network parameters in this work include three parts, canonical correspondence network $F_n$, shape prior network $F_0$, and shape posterior network $F_1$.

\begin{figure}[t]
	\centering
	\includegraphics[width=\linewidth]{figures/pipeline/pose.pdf}%
	\caption{The architecture of our canonical correspondence network. The architecture is modified from \cite{Rempe:etal:NIPS2020}. It extracts global features and spatiotemporal local features from the PointNet encoder~\cite{Qi:etal:CVPR2017} and spatial local features from the PointNet++ encoder~\cite{Qi:etal:NIPS2017}. These features are concatenated and passed to an MLP to regress the canonical shape correspondence and the associated confidence.}
	\label{fig:ch5_nocs_network}

\end{figure}

\cref{fig:ch5_nocs_network} shows the architecture of our canonical correspondence network $F_n$. It takes the partial pointcloud $\generalThree{C_L}{v}{}$ from the depth measurements as input and predicts its correspondence $\mbf v$ in canonical space and the associated confidence $w$.
We train the canonical correspondence network using partial pointclouds generated from the synthetic shapenet dataset~\cite{Shapenet:ARXIV2015}. During the training, we augment the input pointcloud with random object poses and solve the 7DoF object poses using \cref{eq: init loss}. To help network prediction robust to outliers, we also add random depth outliers in the pointcloud generation to learn the correspondence confidence $w$ in a self-supervised way. The solved pose is compared to the augmented ground-truth pose and the whole network is trained end-to-end since the estimation is differentiable. 

\begin{equation}
	\label{eq: init loss}
	\argmin_{s_{O_n}, \T{C_L}{O_n}} \sum_{\pixel{L} \in M_n} w \left( \mbf v - \frac{1}{s_{O_n}}\T{C_L}{O_n}^{-1} \generalThree{C_L}{v}{} (\pixel{L}) \right).
\end{equation}

\begin{figure}[t]
	\centering
	\includegraphics[width=0.8\linewidth]{figures/pipeline/deepsdf.pdf}%
	\caption{The architecture of the shape prior network~\cite{Park:etal:CVPR2019}. The input vector is fed through a decoder, which contains eight fully-connected (FC) layers with one skip connection. FC+ denotes a FC with a following softplus activation and the last FC layer output a single SDF value.}
	\label{fig:ch5_deepsdf}
\end{figure}

\begin{figure}[t]
	\centering
	\includegraphics[width=\linewidth]{figures/pipeline/occupancy.pdf}%
	\caption{The architecture of our shape completion network, modified from CONet~\cite{Peng:etal:ECCV2020}. The encoder extracts the TSDF feature vector $\theta_t[\mbf v] \in \mathbb{R}^{32} $ and the TSDF confidence vector $\theta_c[\mbf v] \in \mathbb{R}^{1}$ from TSDF feature volume and TSDF confidence volume, respectively, and concatenates them with a latent code $z_1$ as an input to the network. It goes through 3 fully-connected (FC) ResNet-blocks to extract local latent features, which are then fed into an occupancy decoder~\cite{Mescheder:etal:CVPR2019} to predict occupancy probabilities on the position vector $\mbf v$. }
	\label{fig:ch5_shape_completion_net}
\end{figure}

We use the pre-trained off-shelf network weights from the category-level shape prior network DeepSDF \cite{Park:etal:CVPR2019} for $F_0$, which was also trained in the synthetic shapenet dataset~\cite{Shapenet:ARXIV2015}. Its architecture is visualized in \cref{fig:ch5_deepsdf}.

\cref{fig:ch5_shape_completion_net} shows the architecture of our posterior shape completion network $F_1$. It takes the input of a TSDF feature volume and a TSDF confidence volume, which are extracted separately from a partial TSDF volume and its weight volume. The input of TSDF confidence volume is designed to balance the observed depth measurement and shape prior information. The unseen part has TSDF weight of zero value and biases towards prior shape and will gradually switch to 3D reconstruction when more depth information is fused into the corresponding TSDF voxel. We additionally concatenate the inputs with a latent code $\mbf z_1 \in \mathbb{R}^{32}$ so that the hidden space can be optimised to generate novel shapes, as shown in \cref{fig:opt_latent}. The shape completion network $F_1$ can predict a complete object geometry represented in a continuous occupancy function by inferring an occupancy value $o$ on any given 3D position $\mbf v \in \mathbb R^3$ in canonical space.

\begin{figure}[t]
	\centering
	\includegraphics[width=\linewidth]{figures/opt_latent/opt_latent-crop.pdf}%
	\caption{Editing the conditioned latent code can change the geometry of the unobserved part in the object model.}
	\label{fig:opt_latent}
\end{figure}

Since the partial observation in reality mostly happens due to self-occlusions and sometimes also due to occlusions from other objects, we rendered depth maps using objects in the shapenet dataset~\cite{Shapenet:ARXIV2015}. To train the posterior shape completion network, we rendered depth maps for each object in the shapenet dataset~\cite{Shapenet:ARXIV2015} to simulate partial depth observations.
We use the occupancy loss defined in \cref{eq: occupied} to encourage the completed shape to be similar to the ground-truth one. 
\begin{equation}
	\label{eq: occupied}
	E_\mathrm{occ} = -\sum_{{\mbf v}} [o_{\mbf v} \log(o_{\mbf v}^*) + (1-o_{\mbf v}) \log(1-o_{\mbf v}^*)].
\end{equation}
Similar to the training in DeepSDF~\cite{Park:etal:CVPR2019}, different object shapes belonging to the same category have different latent codes, but share the same decoder network weight. We make different partial observations of the same object shape share the same latent code.

\bibliographystyle{IEEEtran}
\bibliography{robotvision}